\documentclass[10pt,twocolumn,letterpaper]{article}

\usepackage{cvpr}              

\usepackage[dvipsnames]{xcolor}

\definecolor{cvprblue}{rgb}{0.21,0.49,0.74}
\usepackage[pagebackref,breaklinks,colorlinks,citecolor=cvprblue]{hyperref}
\usepackage{booktabs}
\usepackage{multirow}
\newcommand{\op}[1]{\operatorname{#1}}
\newcommand{\mbf}[1]{\mathbf{#1}}
\usepackage{amsmath}
\usepackage{amsfonts}
\usepackage{float}
\usepackage{hyperref}
\usepackage{indentfirst}
\usepackage[accsupp]{axessibility}
\def\paperID{15166} 
\def\confName{CVPR}
\def\confYear{2024}

\title{Language Model Guided Interpretable Video Action Reasoning}

\author{Ning Wang$^{1*}$, Guangming Zhu$^{1\dag}$\thanks{Ning Wang and Guangming Zhu are co-first authors.}, HS Li$^{1}$, Liang Zhang$^{1}$\thanks{Liang Zhang and Guangming Zhu are both the corresponding authors.}, Syed Afaq Ali Shah$^{2}$, Mohammed Bennamoun$^{3}$\\
    $^{1}$Xidian University, $^{2}$Edith Cowan University, $^{3}$University of Western Australia\\
    {\tt\small \{ningwang, hsli\}@stu.xidian.edu.cn, \quad \{gmzhu, liangzhang\}@xidian.edu.cn,} \\
    {\tt\small afaq.shah@ecu.edu.au, mohammed.bennamoun@uwa.edu.au}
}

\begin{document}
\maketitle
\begin{abstract}
While neural networks have excelled in video action recognition tasks, their "black-box" nature often obscures the understanding of their decision-making processes.
Recent approaches used inherently interpretable models to analyze video actions in a manner akin to human reasoning.
These models, however, usually fall short in performance compared to their “black-box” counterparts.
In this work, we present a new framework named  \textbf{La}nguage-guided \textbf{I}nterpretable \textbf{A}ction \textbf{R}ecognition framework (\textbf{LaIAR}).
LaIAR leverages knowledge from language models to enhance both the recognition capabilities and the interpretability of video models.
In essence, we redefine the problem of understanding video model decisions as a task of aligning video and language models.
Using the logical reasoning captured by the language model, we steer the training of the video model.
This integrated approach not only improves the video model's adaptability to different domains but also boosts its overall performance.
Extensive experiments on two complex video action datasets, Charades \& CAD-120, validates the improved performance and interpretability of our LaIAR framework.
The code of  \textbf{LaIAR} is available at \href{https://github.com/NingWang2049/LaIAR}{https://github.com/NingWang2049/LaIAR}.
\vspace{-0.2cm}
\end{abstract}
\vspace{-0.3cm}    
\section{Introduction}
\label{sec:intro}

Building on the advancements of deep learning in image recognition \cite{krizhevsky2012imagenet, szegedy2015going,he2016deep}, neural network (NN) models have become the leading approach for video-related challenges, including action recognition \cite{plizzari2022e2, sun2022human, lee2023decomposed}.
Yet, many of the top-tier action recognition techniques \cite{li2022improving, wang2023exploring} deploy NNs in an opaque, black-box fashion.
This lack of transparency does not offer clear justification for their decisions, hindering their utility in various real-world contexts \cite{jing2022inaction}, especially those with rigorous security demands.
These considerations drive us to develop an action reasoning system that pairs exceptional performance with clear interpretability.

\begin{figure}[t]
  \centering
  \includegraphics[width=\linewidth]{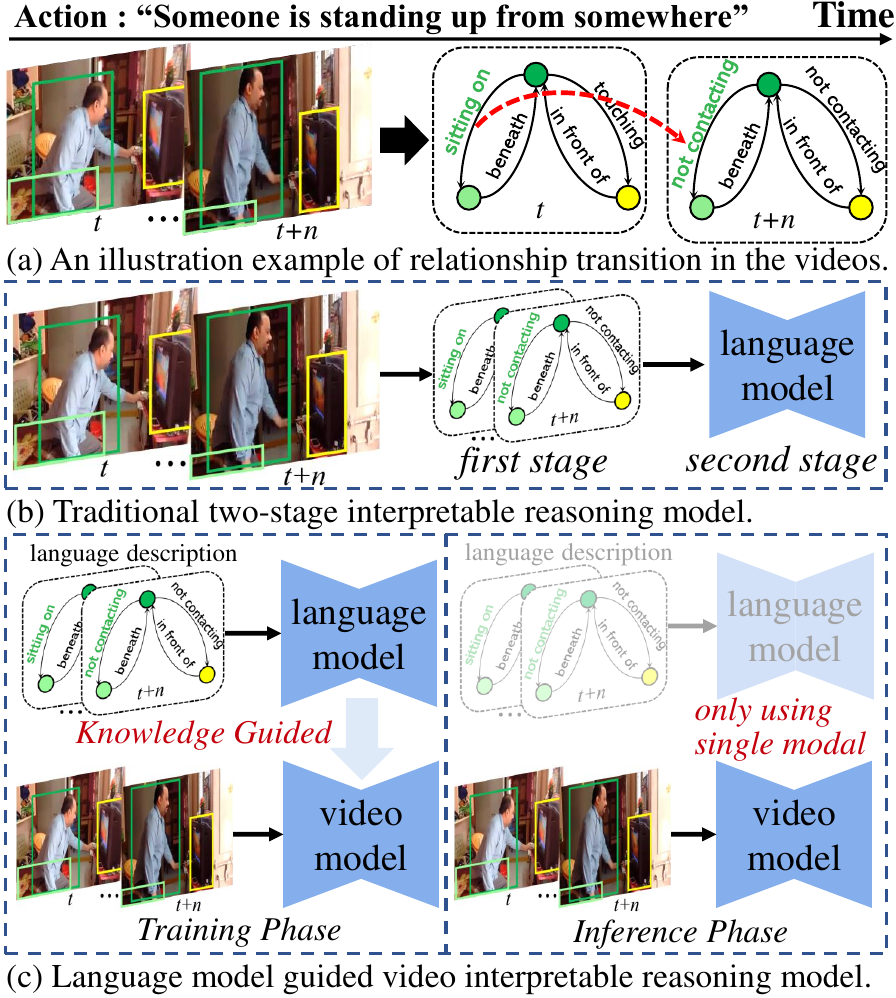}
   \caption{(a) An example of action that can be decomposed into relationship transitions (\textit{i.e.}, when the transition is \textbf{‘sitting on’} $\rightarrow$ \textbf{‘not contacting’} between $<$person, bed$>$ pair, it represents the action "Someone is standing up from somewhere".).
(b) Traditional two-stage methods usually predict the scene graph first, and then use language models to capture the semantic-level relationship transitions.
(c) Our method exploits a language model to guide the video model to capture the relationship transition during training. During inference, our method processes videos and directly recognizes actions, providing supportive evidence.}
   \label{fig:intro}
   \vspace*{-0.7cm}
\end{figure}

Most of the current interpretable action recognition techniques \cite{luo2014learning, soo2017interpretable, meng2019interpretable} aim to elucidate the decision-making process of NNs using \textit{post-hoc} explanations, with a particular emphasis on gradient-based and perturbation-based approaches.
However, despite notable advancements, these explanations can be problematic because they might not be faithful to what the network computes, as highlighted by \cite{rudin2019stop}.
A compelling direction in interpretability revolves around the concept of \textit{built-in} explanation models  \cite{zhuo2019explainable, hua2022towards, jin2022complex, ou2022object}.
The essence of these models is their inherent interpretability right from the design stage.
Recent strategies decompose a complex action into temporal transitions of human-object relationships, drawing inspiration from the event segmentation theory \cite{kurby2008segmentation}.
An illustration in Figure \ref{fig:intro} (a) depicts that for a relationship involving a $<$person, bed$>$ pairing, a transition sequence from \textbf{‘sitting on’} to \textbf{‘not contacting’} signifies the action of "Someone is standing up from somewhere".
This methodology facilitates action recognition by pinpointing semantic transitions through language models, offering a granular insight into action execution.
As shown in Figure \ref{fig:intro} (b), Jin and Ou et al. \cite{jin2022complex, ou2022object} extract spatio-temporal scene graphs from video  content and apply Markov Logic Network (MLN) based probabilistic logical inference and relation reasoning graphs to create an interpretable representation for a variety of complex actions, respectively.
\textit{\textbf{However, it is believed that such models will perform worse than their black-box alternatives}} \cite{gunning2019darpa}.
Moreover, these methods divide the process into two stages, namely scene graph prediction and relation modeling.
Optimizing these components separately might lead to sub-optimal results.
In this paper, we propose to harness the explicit logical inference rules of an interpretable language model to guide the learning process of a video black-box model. 
Interpretability and strong performance can be attained by focusing solely on the video model during the inference stage. 
To achieve this, two main challenges arise: 1) Designing a language model that can automatically grasp logical reasoning patterns, sidestepping manual rule creation. 2) Developing a decoupled language-video model architecture that enables the language model to guide the video model's training process.

To address the aforementioned challenges, we have developed a new framework called \textbf{La}nguage-guided \textbf{I}nterpretable \textbf{A}ction \textbf{R}ecognition framework named \textbf{LaIAR}.
As depicted in Figure \ref{fig:intro} (c), LaIAR constructs an action recognition model that both implicitly and explicitly exploits fine-grained knowledge of relationship transitions from an interpretable built-in language model.
Specifically, we use dynamic token transformers (DT-Former) to both video and language inputs, selectively focusing on important relationships in a data-driven manner, and disregarding non-contributory ones.
We aim to redefine the traditional decision interpretation challenge of video models towards a visual-language relation alignment problem.
\textit{\textbf{The relationship prioritization determined by the language model then explicitly guides the video model in identifying the most relevant relationships}}.
We propose a learning strategy to facilitate knowledge transfer between language and video to improve the performance of the video model.
A key feature of \textbf{LaIAR} is its modular design: during the inference phase, only RGB data serves as input to predict actions, providing a direct and transparent justification.

To summarize, our contributions are three-fold: 
1) We propose a novel \textbf{LaIAR} framework that can automatically mine fine-grained relation transitions from data and create interpretable representations for various complex actions.
2) We design a decoupled cross-modal knowledge transfer architecture that leverages useful knowledge from language models to improve the performance and interpretability of the  video model at training time, and achieves high-performance interpretable reasoning for videos at test time.
3) Our method achieves state-of-the-art results on two large-scale action recognition benchmarks.
\section{Related Work}
\label{sec:related}

\subsection{Interpretable Video Action Recognition}
Interpretable video action recognition methods can be categorized into two types: \textit{post-hoc} method and \textit{built-in} method.
\textit{post-hoc} techniques generate explanations for the network’s decision-making process after the network is trained.
\cite{meng2019interpretable} introduced an interpretable and easy plug-in spatial-temporal attention mechanism for video action recognition to improve the interpretability of the model for video action recognition.
\cite{soo2017interpretable} developed an interpretable temporal convolutional network to explain the decision-making process of action recognition through each of the learned filters in a Res-TCN.
\cite{luo2014learning} combined both global dynamics and local details to learn human action, using gradient-weighted class activation mapping (Grad-CAM) to visualize the model’s attention to action-critical regions.
Although these methods have the advantage of not imposing any model constraints, they may be incomplete or unfaithful to the model's reasoning \cite{rudin2019stop}. 
In contrast, \textit{built-in} methods restrict the interpretation to be consistent with the model's inferences.
\cite{zhuo2019explainable} approached action reasoning by modeling semantic-level state transitions between two consecutive frames as defined by domain experts.
In \cite{hua2022towards}, a method is proposed to achieve interpretability of action recognition by incorporating qualitative spatial reasoning and extracting salient relation chains. 
Some recent methods, like \cite{jin2022complex, ou2022object} decompose complex action into continuous relationship transitions according to the event segmentation theory \cite{kurby2008segmentation}. These methods model the relationship transitions at the semantic level to recognize actions.
In this paper, we propose to construct a high performance interpretable-by-design action classifier by guiding a video model with an interpretable language model.

\subsection{Adaptive Inference in Transformers.}
As their popularity soars, adaptive inference for language and vision transformers has caught the attention of researchers.
In \cite{ye2021tr}, an adaptive language transformer is proposed to achieve fixed-scale reduction of the input sequence to improve inference speed by dynamically selecting important tokens and removing the irrelevant ones. 
In \cite{modarressi2022adapler}, a threshold mechanism is introduced to determine the importance of each token and dynamically select the tokens according to the importance of the input sequence. 
\cite{kim2022learned} used the mean of attention matrix column values of the transformers to determine the importance score of each token, facilitating token pruning.
\cite{yin2022vit} developed an adaptive token generation mechanism to determine the required number and size of tokens, thereby reducing the computational and memory overhead of the model on images.
\cite{wang2022efficient} devised a token selection framework to dynamically select important tokens across the temporal and spatial dimensions of the video input.
In this paper, we propose a lightweight token selection method based on Gumbel-Softmax and apply it to our cross-modal transformers for spatio-temporal token selection.

\subsection{Cross-modal Knowledge Transferring}
The past few years have seen an increasing interest in cross-modal knowledge transfer techniques for detection and segmentation tasks.
\cite{lin2021effects} introduced a method to transfer the motion-related knowledge of unlabeled videos to Human-Object Interactions (HOI) detection to infer rare or unseen HOIs.
In \cite{zhang2022audio}, reliable domain-invariant sound cues are exploited to help video activity recognition models adapt to video distribution shifts. 
Lately, knowledge distillation techniques have been extended to transfer knowledge across different modalities.
For instance, \cite{lee2023decomposed} proposed a decomposed cross-modal distillation framework to improve RGB-based temporal action detection by transferring knowledge from the optical flow modality.
Similarly, \cite{yuan2022x} proposed a modified knowledge distillation method that boosts the performance of single-modal 3D captioning by transferring color and texture-aware information from 2D images into 3D object representations.
In contrast to these methods, we propose a well-designed knowledge-guided framework to enable cross-modal learning by decoupling information transfer in video and language.
\section{The Proposed Approach}

\begin{figure*}[htb]
  \centering
  \includegraphics[width=0.9\linewidth]{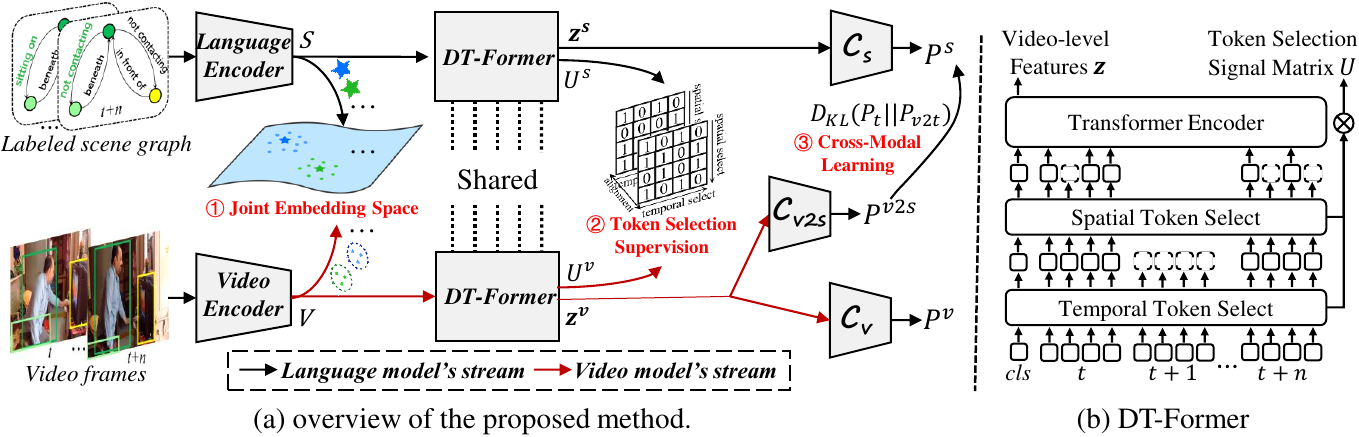}
   \caption{Overview of our \textbf{LaIAR}.
   The architecture comprises a language model (top) which takes the language description (represented as a spatio-temporal scene graph in \cite{ji2020action}) as input and a video model (bottom) which takes the video frames as input.
   Both models use DT-Former to capture key relational transitions to recognize actions.
   We transfer knowledge across modalities using a learning scheme (\textit{i.e.}, Joint Embedding Space, Token Selection Supervision and Cross-Modal Learning), which can help video model benefit from language model during training.
   For inference,  only the video model is considered.
   }
   \label{fig:method}
   \vspace*{-0.4cm}
\end{figure*}

Our proposed method is designed to exploit multi-modality by enabling information transfer from language descriptions to videos.
This enables the video model to effectively learn from the language model effectively.
This is achieved by the video model mimicking the output of the language model, thereby leveraging the intrinsic capabilities of the language model.
Specifically, the video frames and the language description (represented as a spatio-temporal scene graph in \cite{ji2020action}) are \textbf{\textit{first}} processed by the encoding network to extract the paired visual and semantic relationship representations.
\textbf{\textit{Then}}, these paired visual and semantic relationship representations are separately fed to DT-Former module, which models the key relationship transition for action recognition.
\textbf{\textit{Finally}}, we propose a learning scheme (\textit{i.e.}, Joint Embedding Space, Token Selection Supervision and Cross-Modal Learning) to improve the performance and interpretability of the video model by facilitating the knowledge transfer from the language model to the video model.
An overview of our proposed method is shown in Figure \ref{fig:method} (a).
Note that, in our proposed approach, only the video model is employed for inference once training is complete.

\subsection{Architecture}
\subsubsection{Video and Language Encoder}
Given a video consisting of $T$ frames with $N$ entities of either human or object classes, we use Faster R-CNN \cite{ren2015faster} with ResNet-101 \cite{krizhevsky2012imagenet} backbone to detect these entities and extract their features from the video.
For the frame $I_t$ at time step $t$, the visual features $\{v_{(t,1)}, v_{(t,2)}, ..., v_{(t,N)}\}$, bounding boxes $\{b_{(t,1)}, b_{(t,2)}, ..., b_{(t,N)}\}$ and object category $\{c_{(t,1)}, c_{(t,2)}, ..., c_{(t,N)}\}$ of the objects proposals are supplied by the detector.
Between each $<$human, object$>$ pair in the frame, there is a set of relationships $R_t = \{r_{(t,1)}, r_{(t,2)}, ..., r_{(t,K)}\}$.
Concatenating the visual appearance, spatial information and category embedding between the $i$-th human and $j$-th object proposals can represent the visual relation feature $\mathbf{v}_{(t,k)}$,  as follows:
\begin{equation}
  \mathbf{v}_{(t,k)} = \left[ W_s v_{(t,i)}, W_o v_{(t,j)},  W_u \varphi (u_{(t,ij)} \oplus f_{box}(b_{(t,i)}, b_{(t,j)})) \right]
\end{equation}
where $W_s$, $W_o$ and $W_u$ represent the parameter matrix of the linear transformation.
$\left[,\right]$ is concatenation operation, $\varphi$ is flattening operation and $\oplus$ is element-wise addition.
$u_{(t,ij)}$ the visual feature of the union box of $b_{(t,i)}$ and $b_{(t,j)}$ extracted from the detector.
$f_{box}$ maps the 2-channel binary spatial configuration map of bounding boxes $b_{(t,i)}$ and $b_{(t,j)}$ into features of the same dimension as $u_{(t,ij)}$.

Unlike the visual relation feature, the semantic relation feature $\mathbf{s}_{(t,k)}$ provides high-level descriptions of the relationship between humans and objects in the videos.
\textit{The visual relationship categories are either provided as ground-truth or determined by the fine-tuned visual relationship detection network \cite{cong2021spatial}}.
The features of the semantic relation are obtained by concatenating the three features as follows:
\begin{equation}
  \mathbf{s}_{(t,k)} = \left[s_{(t,i)}, r_{(t,ij)} , s_{(t,j)} \right]
\end{equation}
where the $r_{(t,ij)}$ is extracted by embedding the visual relationship category to the semantic feature space.
The category embedding vectors $s_{(t,i)}$ and $s_{(t,j)}$ are determined by the categories of human and object, respectively.

Given the features $\{\mathbf{v}_{(t,k)}\}_{t=1,k=1}^{T,K}$ and the features $\{\mathbf{s}_{(t,k)}\}_{t=1,k=1}^{T,K}$, we further map the visual relation feature and the semantic relation feature into a joint embedding space as follows:
\begin{align}
    \mathcal{V} = {f}_{\textsc{v}}\Big( \{\mathbf{v}_{(t,k)}\}_{t=1,k=1}^{T,K} \Big), \quad
    \mathcal{S} = {f}_{\textsc{s}} \Big( \{ \{\mathbf{s}_{(t,k)}\}_{t=1,k=1}^{T,K} \Big).  \nonumber
\end{align}
where the ${f}_{\textsc{v}}$ and ${f}_{\textsc{s}}$ are the visual encoder and the semantic encoder, respectively.
In each encoder, each element of the input is first mapped to a local representation via a linear projection. 
We then apply a Generalized Pooling Operator (GPO) \cite{chen2021vseinfty} to aggregate the input, creating a global representation. 
This global representation is combined with each local representation along the channel dimension. This approach helps to use the contextual information from the entire sequence.
The visual embedding $\mathcal{V} \in \mathbb R^{T \times K \times D}$ and semantic embedding $\mathcal{S} \in \mathbb R^{T \times K \times D}$ are aligned in the  spatio-temporal dimension, where $D$ denotes the dimension in the common space.

\subsubsection{Dynamic Token Transformers}
Video understanding shares several high-level similarities with natural language processing (NLP), as they are both fundamentally based on sequential structures \cite{bertasius2021space}.
Our intuition is that we can easily model visual and semantic relations simultaneously from joint embedding space (see Sec \ref{vsjes} for details).
Therefore, we introduce a shared dynamic token transformers (DT-Former), which employs the transformer structure to capture key relationship transitions for action reasoning. 
It mainly consists of the adaptive token selection and the video transformer module.
The adaptive token selection module calculates the contribution score of each token to the classification output, and tokens with lower contribution scores will be discarded.
The retained tokens, \textit{i.e.} important relationship representations, are fed to the video transformer module to capture relation transition cues.
Figure \ref{fig:method} (b) shows the architecture of DT-Former.
Since the video model and the language model share the same DT-Former, we denote the input as $X = \{x_{(t,k)}\}_{t=1,k=1}^{T,K}$ for simplicity.
We add a learnable spatiotemporal positional embedding $e_{(t,k)}^{pos}$ to each vector $x_{(t,k)}$ to obtain the embedding token $x_{(t,k)}^{(0)}$.
The superscript corresponds to the layer of the transformer encoder.


\textit{\textbf{Adaptive Token Selection.}}
Following the ViT approach \cite{dosovitskiy2020image}, we concatenate a special learnable vector (${x}_{(0,0)}^{(0)}$=$x_{class}$) representing the embedding of the \verb|[class]| token in the first position of the sequence.
As a large number of relationships between human and objects in a scene are usually redundant, it is essential to reduce these relationships.
Inspired by the recent work on token reduction for accelerating transformer inference, we formulate parsing important relations as a token selection problem.
To determine whether a token is discarded or retained, we introduce a \textit{token selector} that consists of an MLP $\sigma$ and a differentiable discrete-valued estimator using the Gumbel-Softmax (GSM) operator:
\begin{equation}
 u_{(t,k)} = \op{GSM}\{ \sigma ([W_1 x_{(t,k)}^{(0)}, W_2 x_{class} ]) \}
 \label{gsm}
\end{equation}
where $W_1$ and $W_2$ represent the linear matrices for dimension compression. 
We concatenate \verb|[class]| tokens that represent the global representation with input tokens to exploit contextual information of the entire sequence.
The binary output $u_{(t,k)} = 0$ indicates that the $t$-th token of frame $t$ is to be discarded and $u_{(t,k)} = 1$ is to be retained.
Token selection can be represented as: $y_{(t,k)}^{(0)} = u_{(t,k)} x_{(t,k)}^{(0)}$.
Note that this operation is differentiable, allowing for end-to-end training tailored for token selection.

To ensure consistent token reduction across consecutive frames, we apply token selector to both the temporal and spatial dimensions.
Recent efforts in the field of frame sampling \cite{zhi2021mgsampler} indicate inherent temporal redundancy in frames. 
Inspired by this, we first focus on salient frames over the entire time horizon, and then delve into those frames to find key relationships.
For the input tokens $X = \{x_{(t,k)}^{(0)}\}_{t=1,k=1}^{T,K}$, we \textit{\textbf{first}} apply an average-pooling operation to tokens in the spatial dimension to get a sequence of temporal-based tokens $\{x_{(t)}^{(0)}\}_{t=1}^{T}$, and \textit{\textbf{then}} feed it to the \textit{token selector} to generate temporal-based selection signal matrix $\{\grave{u}_{(t)}\}_{t=1}^{T}$.
\textit{\textbf{Finally}}, we repeat it along the spatial dimension to obtain $\grave{U} = \{\grave{u}_{(t, k)}\}_{t=1, k=1}^{T, K}$ for downstream processing.
Similarly, we perform the \textit{token selector} on each frame separately to generate a selection signal matrix. The spatial-based selection signal matrix can be expressed as $\acute{U} = \{\acute{u}_{(t, k)}\}_{t=1, k=1}^{T, K}$. 
Further, the final selection signal matrix can be expressed as $U = \grave{U} \cdot \acute{U} = \{u_{(t, k)}\}_{t=1, k=1}^{T, K}$.
We use matrix multiplication for token selection: $Y=U \cdot X = \{y_{(t,k)}^{(0)}\}_{t=1,k=1}^{T,K}$.


\textit{\textbf{Transformer Encoder.}}
In order to model the relationship transition in videos, the token $Y=\{y_{(t,k)}^{(0)}\}_{t=1,k=1}^{T,K}$ are fed to stack of transformer blocks which compute the spatial and temporal self-attention jointly. 
We convert $Y$ into a set of sequences $\mbf{Y}^{(0)}=\{y_{(p)}^{(0)}\}_{p=1}^{T \times K}$, which are then fed into the transformer encoder to extract a video-level representation:
\begin{align}
    \mbf{Y^\prime}^\ell &= \op{MSA}(\op{LN}(\mbf{Y}^{\ell-1})) + \mbf{Y}^{\ell-1} \\
    \mbf{Y}^\ell &= \op{MLP}(\op{LN}(\mbf{Y}^{\ell})) + \mbf{Y^\prime}^{\ell} \\
    \mbf{z} &= \op{LN}(\mbf{Y}_{(0)}^L) \label{eq:final_rep}
\end{align}
where MSA() and LN() denotes multiheaded self-attention and LayerNorm \cite{ba2016layer}, respectively.
$L$ represents the number of transformer blocks.
$\mbf{z}$ denotes the video-level representation, which can be used to predict the final action classes.
Based on the above steps, we can get the selection signal matrix $U^v = \{u_{(t, k)}^v\}_{t=1, k=1}^{T, K}$, video-level representation $\mbf{z^v}$ of the video model, the selection signal matrix $U^s = \{u_{(t, k)}^s\}_{t=1, k=1}^{T, K}$ and video-level representation $\mbf{z^s}$ of the language model.

\subsubsection{Classification Head}
The two classification heads of video and language model predict, $P^{v} = \{p_{(c)}^{v}\}_{c=1}^{C}$ and $P^{s} = \{p_{(c)}^{s}\}_{c=1}^{C}$ respectively for each branch, where $C$ is the number of classes.
We minimize the cross-entropy losses between action scores $P^{v}$, and $P^{s}$ and the ground-truth action labels for each action category, denoted as $\mathcal{L}_{v}$ and $\mathcal{L}_{s}$, respectively.
The overall loss of two branches can be written as:
\begin{equation}
  \mathcal{L}_{cls} = \mathcal{L}_{v} + \mathcal{L}_{s}
\end{equation}

The two classification heads only utilize the private information of each modality, in order to allow the video model to benefit from the knowledge of the language model, we propose an additional classification head to estimate the other modality’s output: the video model estimates the language model ($P^{v2s} = \{p_{(c)}^{v2s}\}_{c=1}^{C}$).
By mimicking not only the class with maximum probability, but also the whole distribution, more information is exchanged, leading to softer labels, which is more beneficial for training our model.

\subsection{Learning Scheme}

The goal of our learning scheme is to transfer information across modalities in a controlled manner thus allowing the video model to learn from the language model.
This auxiliary objective can effectively improve the performance of the video modality and does not require additional labels from the datasets.
Here, we define the visual-semantic joint embedding learning, our token selection supervision loss and an additional cross-modal learning method.

\subsubsection{Visual-Semantic Joint Embedding Space}
\label{vsjes}
Our approach begins by aligning the visual and semantic relation representations within a shared vector space. In this configuration, each visual embedding $\hat{\mathbf{v}}_{(t,k)} \in \mathcal{V}$ and $\hat{\mathbf{s}}_{(t,k)} \in \mathcal{S}$ pair converge to proximate points.
This visual-semantic joint embedding has two main advantages: 1) it helps the video model to improve its generalization since semantic representations are invariant to complex appearance variations. 2) it enables the video model to explicitly represent the relationship transition process, since the visual-semantic joint embedding space can provide semantic labels for each visual relation representation.
In this paper, we introduce the contrastive learning of visual-semantic joint embedding and discuss its implementation as following.

Given a mini-batch $B = \{(\hat{\mathbf{v}}_{(0,0)}, \hat{\mathbf{s}}_{(0,0)}), ... \}$ of visual-semantic relationship representation pairs, the contrastive learning objective encourages embeddings of positive pairs $(\hat{\mathbf{v}}_{(t,k)}, \hat{\mathbf{s}}_{(t,k)})$ to align with each other, while pushing embeddings of the negative pairs apart.
Formally, the contrastive loss $\mathcal{L}_{sim}$ is formulated using the symmetric contrastive loss, as follows:
\begin{equation}
L_{sim} = -\frac 1 {2|\mathcal{B}|} \sum_{i=1}^{|\mathcal{B}|} \left(
\overbrace{\log \frac {e^{t\mathbf{x}_i \cdot \mathbf{y}_i}} {\sum_{j=1}^{|\mathcal{B}|} e^{t\mathbf{x}_i \cdot \mathbf{y}_j}}}^\text{visual$\rightarrow$semantic} +
\overbrace{\log \frac {e^{t\mathbf{x}_i \cdot \mathbf{y}_i}} {\sum_{j=1}^{|\mathcal{B}|} e^{t\mathbf{x}_j \cdot \mathbf{y}_i}}}^\text{semantic$\rightarrow$visual} \right)
\end{equation}
where $\mathbf{x}_i=\frac {\mathbf{v}_{(t,k)}} {\|\mathbf{v}_{(t,k)}\|_2}$ and $\mathbf{y}_i=\frac {\mathbf{s}_{(t,k)}} {\|\mathbf{s}_{(t,k)}\|_2}$.
$|\mathcal{B}|$ is size of the mini-batch $B$. $t$ is the learnable temperature parameter.

\subsubsection{Token Selection Supervision}
We aim that the key relations obtained by the language model can guide the video model to perform key relations mining.
Therefore, we align the token selection signal matrix of the two models.
We minimize the mean-squared loss between the token selection signal matrix $u_{(t,k)}^v$ of the video model and the token selection signal matrix $u_{(t,k)}^s$ of the language model.
To maintain the token selection signal matrix’s sparsity, we apply L1 norm to allow $u_{(t,k)}^s$ to have a small number of non-zero values.
The token selection supervision loss is defined as:
\begin{equation}
  \mathcal{L}_{tss} = \frac{1}{T}\sum\limits_{t=1}^T\frac{1}{K}\sum\limits_{k=1}^K\left(\|u_{(t,k)}^v-u_{(t,k)}^s\|+\|u_{(t,k)}^s\|_1\right)
\end{equation}

\subsubsection{Cross-Modal Learning}
Given that the visual modality is highly sensitive and the semantic modality more robust to the domain shift, the robust semantic modality can guide the sensitive visual modality to the correct classification.
We allow the video model estimate the entire distribution of the language model's prediction. 
Through cross-modal learning, we aim to transfer knowledge from the language model to the video model.
We choose KL divergence for the cross-modal loss $\mathcal{L}_{\mathrm{xm}}$ and define it as follows:
\begin{equation}
\mathcal{L}_{\mathrm{xm}}=D_{\mathrm{KL}}(P^{s}||P^{v2s})  \\
=-\sum\limits_{c=1}^C p_{(c)}^{s}\log\frac{p_{(c)}^{s}}{p_{(c)}^{v2s}}
\end{equation}

\subsection{Training and Inference}

\subsubsection{Training}
During the training process, we adopt a random sampling strategy to sample fixed $T$ frames for each video.
In order to obtain the best possible performance, our framework jointly trains the classification objective and the learning scheme in an end-to-end manner.
The final loss is:
\begin{equation}
  \mathcal{L} = \mathcal{L}_{cls} + \delta \mathcal{L}_{sim} + \zeta 
  \mathcal{L}_{tss} + \eta \mathcal{L}_{xm} 
\end{equation}
where $\delta$, $\zeta$ and $\eta$ are hyperparameters that control the importance of the learning scheme. We use $\delta=0.1$, $\zeta=1$ and $\eta=0.1$ in our experiments.

\subsubsection{Inference}
During the inference process, a uniform sampling strategy is applied to sample fixed $T$ frames for each video.
Only the video model is considered for inference.
The explanation of the reasoning process can be explicitly shown by the proximity of the visual representation to the semantic representation in the joint embedding space.
\section{Experiments}

\begin{table}[htb]
\caption{Ablation study on the Charades and CAD-120 datasets using each proposed module.
"S" denotes the spatial token selection and "T" denotes the temporal token selection.
}
\vspace{-3mm}
\centering
\scalebox{0.9}{
\begin{tabular}{@{}l|cc|cc@{}}
\toprule
\multirow{2}{*}{Methods} & \multicolumn{2}{c|}{Charades} & \multicolumn{2}{c}{CAD-120} \\ & mAP (\%)$\uparrow$ & Num $\downarrow$ & mAR $\uparrow$ & Num $\downarrow$ \\ \midrule
DT-Former w/o S,T & 61.4 & 36.6 & 0.73 & 49.8 \\
DT-Former w/o S & 61.2 & 31.1 & 0.72 & 17.4 \\
DT-Former w/o T & 61.2 & 30.7 & 0.74 & 16.7 \\
DT-Former & 61.1 & 26.0 & 0.75 & 14.2 \\ \bottomrule
\end{tabular}}
\label{component}
\end{table}

\subsection{Datasets and Metrics}
\textbf{Datasets.} 
We conduct our experiments on two extensive video datasets, detailed as follows:
(1) \textbf{Charades \cite{sigurdsson2016hollywood}}. 
It contains 157 action classes and consists of about 9.8k untrimmed videos, among which 7.9k are used for training and 1.8k for testing.
Each video contains an average of 6.8 distinct action categories and multiple actions can happen at the same time, which makes the recognition extremely challenging. 
The Action Genome dataset \cite{ji2020action} provides fine-grained annotations for the Charades dataset, which provides frame-level relation annotations for videos. Overall, it annotates 476K object bounding boxes and 1.72M relations.
(2) \textbf{CAD-120}. 
Introduced by \cite{koppula2013learning}, the CAD-120 dataset is an RGB-D dataset designed for activity understanding. It contains 551 video clips of 4 subjects performing 10 different activities in different environments, such as a kitchen, a living room, and office, etc.
To train on our method, we leverage the re-annnotated version provided by \cite{zhuo2019explainable}, which provides detailed relationships and attributes for the video frames.

\textbf{Evaluation protocol.}
Following the experimental protocol of \cite{ji2020action}, We measure multi-label action recognition performance in term of the Mean Average Precision (mAP) on Charades dataset.
For CAD-120 dataset, we calculate the Mean Average Recall (mAR) to evaluate whether the model successfully recognizes the performed actions.

\subsection{Ablation Studies}

\subsubsection{Effectiveness of Each Module}
Table \ref{component} reports the effectiveness of each module of the proposed architecture. We evaluate the performance of the DT-Former using different settings, \textit{i.e.}, canceling spatial token selection or canceling temporal token selection or both.
Here, the DT-Former corresponds to the performance obtained by the video model.
The metric scheme 'Num' refers to the average number of tokens retained per video after token selection.
On the CAD-120 dataset, we noted a modest improvement in the accuracy of our architecture, attributable to the spatial-temporal token selection. The CAD-120 dataset typically features videos with a single action, usually characterized by a pair of relational transitions. By eliminating irrelevant features, the model's risk of overfitting is reduced, thereby enhancing its ability to generalize.
As expected, more tokens are discarded in the CAD-120 dataset than in the complex Charades dataset.
Overall, adding either or both of the token selection modules can reduce the number of tokens without significantly affecting that the recognition performance.
This demonstrates our architecture's capability in identifying key relational transitions for action recognition and disregarding superfluous/redundant relations.

\begin{table}[htb]
\caption{Ablation study of learning scheme on the Charades and CAD-120 datasets. $\checkmark$ indicates that the component is applied in the experiments. }
\vspace{-3mm}
\centering
\scalebox{0.85}{\begin{tabular}{@{}ccc|cc@{}}
\toprule
\begin{tabular}[c]{@{}c@{}}w/\\ $\mathcal{L}_{sim}$ \end{tabular} & \begin{tabular}[c]{@{}c@{}}w/\\ $\mathcal{L}_{tss}$ \end{tabular} & \begin{tabular}[c]{@{}c@{}}w/\\ $\mathcal{L}_{xm}$\end{tabular} & \begin{tabular}[c]{@{}c@{}}mAP on\\ Charades (\%) $\uparrow$ \end{tabular} & \begin{tabular}[c]{@{}c@{}}mAR on\\ CAD-120 $\uparrow$ \end{tabular} \\ \midrule
 &  &  & 61.1 & 0.75 \\
$\checkmark$ & - & - & 62.2 & 0.79 \\
- & $\checkmark$ & - & 61.7 & 0.79 \\
- & - & $\checkmark$ & 62.9 & 0.81 \\
$\checkmark$ & - & $\checkmark$ & 63.4 & 0.83 \\
$\checkmark$ & $\checkmark$ & $\checkmark$ & 63.6 & 0.85 \\ \bottomrule
\end{tabular}}
\label{scheme}
\vspace*{-0.5cm}
\end{table}

\subsubsection{Effectiveness of Learning Scheme}
To explore the effectiveness of the visual-semantic joint embedding ($\mathcal{L}_{sim}$), token selection supervision ($\mathcal{L}_{tss}$) and cross-modal learning ($\mathcal{L}_{xm}$) in our learning scheme, we conduct related ablation studies using different settings, \textit{i.e.}, cancelling one or any two or all our key modules.
As reported in Table \ref{scheme}, the visual-semantic joint embedding, token selection supervision and cross-modal learning components collectively or individually contribute to the final performance improvement.
This improvement confirms our hypothesis that language models can help video models to improve performance through knowledge transfer.
It can be noted that due to the advantages of the learning scheme, the visual model improves from 61.1 to 63.6  in terms of mAP metric and from 0.75 to 0.85  in terms of mAR metric on Charade and CAD-120, respectively.
These results demonstrates that the learning scheme plays an important role in our proposed \textbf{LaIAR}.

\subsubsection{Effectiveness of Robustness Against Domain Shift}
The performance of RGB-based methods drops drastically when the training and testing data do not share the same distribution caused by change of \textbf{scene}, \textbf{camera viewpoint} or \textbf{actor} \cite{zhang2022audio}.
Our proposed model can adapt to video distribution shifts with the aid of  semantic modality, which are invariant to complex appearance variations.
To demonstrate the robustness of our proposed framework to domain shift, we split the Charades dataset into five subsets with non-overlapping training scenes and test scenes.
Table \ref{domain} reports the average and variance of five accuracies for these five subsets.
The variance of our method is clearly stable and indicates robustness to domain shift.


\begin{table}[tb]
\begin{minipage}[r]{0.25\textwidth}
\caption{Ablation study of the domain shift on the Charades dataset.}
\vspace{-3mm}
\centering
\scalebox{0.8}{
\begin{tabular}{c|cc} \hline
& \multicolumn{2}{c}{Accuracy} \\
Method & Average & Variance \\ \hline
STIGPN \cite{wang2023exploring} & 54.1 & 0.30 \\
Ours & 57.2 & 0.11 \\ \hline
\end{tabular}
\label{domain}
}
\end{minipage}
\hspace{1.5mm}
\begin{minipage}[r]{0.2\textwidth}
\caption{Comparison of accuracy using predicted and annotated relationships.}
\vspace{-3mm}
\centering
\scalebox{0.8}{
\begin{tabular}{c|c} \hline
\begin{tabular}[c]{@{}c@{}}Evaluation Mode\end{tabular} & mAP \\ \hline
Prediction & $62.4$ \\
Label & $63.6$ \\ \hline
\end{tabular}
}
\label{scene}
\end{minipage}
\end{table}

\subsubsection{Effectiveness of Using Predicted Relationships.}
As previously stated, visual relationship categories can be manually annotated or identified by the visual relationship detection network \cite{cong2021spatial}.
To explore the impact of the two modes on accuracy, we compared the proposed method based on the ground truth and the predicted semantic relationships during training.
The results, as detailed in Table \ref{scene}, reveal that using predicted semantic relations can indeed enhance the accuracy of the video model on the Charades dataset (achieving 62.4\% mean Average Precision (mAP) versus 61.1\% mAP shown in Table \ref{scheme}). 
Moreover, the accuracy does not significantly decrease when using predicted data instead of ground truth.
This demonstrates the effectiveness of the proposed method in mining relational transformations from real video data.

\subsection{Comparison to the State-of-the-Art}
We compare the action recognition accuracy of the proposed method and the state-of-the-art methods (SoTA) on the Charades and the CAD-120 datasets, respectively.
Table \ref{charades} summarizes the results on Charades. 
It can be seen that our proposed method outperforms several previous 3D CNN approaches in terms of mAP, including I3D \cite{carreira2017quo}, SlowFast \cite{feichtenhofer2019slowfast} and LFB \cite{wu2019long}.
This demonstrates that our method can fully capture action cues through the visual relationship transitions, based on the human/objects information detected from a single video frame (rather than using the entire scene like I3D).
STRG \cite{wang2018videos} and SGFB \cite{ji2020action} model the action based objects and visual relationships, respectively, and overlook explicit modeling of temporal dynamics of the interaction between objects.
Though VideoLN \cite{jin2022complex} and OR2G \cite{ou2022object} takes visual  relationship transitions into account, it is difficult for these methods to achieve accurate action inference due to the limitations of scene graph predictors at test time.
\textit{For comparison in a modality with only RGB video frames, our method achieved the best performance compared with the existing methods.} 
We also evaluated our method in Oracle evaluation mode, which leverages the ground-truth of bounding box and relationships on a frame.
As reported in the last row of Table \ref{charades}, our method still achieves best performance on the Charades dataset.
\textit{It is important to mention that OR2G \cite{ou2022object} used ground-truth scene graphs to enhance its final performance, whereas our network uses only the bounding boxes of humans and objects.}
Despite this, our method demonstrates strong performance in both evaluation modes, validating the effectiveness of our proposed approach.
\begin{table}[htb]
\centering
\caption{Multi-label action recognition performance comparison on the Charades's validation set in term of mAP. SG: ground truth scene graph. Bbox: Bounding Box. Higher values are better.}
\vspace{-3mm}
\scalebox{0.9}{\begin{tabular}{@{}llll@{}}
\toprule
Methods & Backbone & Modality & mAP \\ \midrule
I3D \cite{carreira2017quo} & R101-I3D &  RGB& 15.6\\
VideoMLN \cite{jin2022complex} & R101 & RGB & 38.4 \\ 
STRG \cite{wang2018videos} & R101-I3D-NL & RGB& 39.7\\ 
SlowFast \cite{feichtenhofer2019slowfast} & R101 & RGB& 42.1\\ 
LFB \cite{wu2019long} & R101-I3D-NL & RGB & 42.5\\ 
SGFB \cite{ji2020action} & R101-I3D-NL & RGB & 44.3\\
OR2G \cite{ou2022object} & R101-I3D-NL & RGB & 44.9 \\
\textbf{Ours} & R101-I3D-NL & RGB & \textbf{45.1} \\
\midrule
SGFB Oracle \cite{ji2020action} & R101-I3D-NL & RGB+SG & 60.3 \\
VideoMLN Oracle \cite{jin2022complex} & R101 & RGB+SG & 62.8 \\ 
OR2G Oracle \cite{ou2022object} & R101 & RGB+SG & 63.3 \\
\textbf{Ours Oracle} & R101 & RGB+Bbox & \textbf{63.6} \\
\bottomrule
\end{tabular}}
\label{charades}
\end{table}

For CAD-120 dataset, we follow the same experimental protocol as in \cite{zhuo2019explainable} and divide the long video sequences into small segments based on individual sub-actions and evaluate the average recall metric for each sub-action.
Table \ref{cad120} summarizes the results on CAD-120. 
Explainable AAR-RAR \cite{zhuo2019explainable} interpret the action reasoning process through the changes of relationship between objects or the attribute of objects across time.
Our method is able to give the same explanation and outperforms these methods, achieving state-of-the-art performance with 0.85 mAR.

\begin{table}[htb]
\centering
\caption{Experimental results on CAD-120 for action recognition.}
\vspace{-3mm}
\begin{tabular}{@{}ccc@{}}
\toprule
Methods & Modality & mAR \\ \midrule
\multirow{3}{*}{Temporal Segment \cite{wang2018temporal}} & RGB & 0.42 \\
 & Flow & 0.71 \\
 & RGB + Flow & 0.77 \\ \midrule
Explainable AAR-RAR \cite{zhuo2019explainable} & RGB & 0.80 \\
VideoMLN \cite{jin2022complex} & RGB & 0.83 \\
\textbf{Ours} & RGB & \textbf{0.85} \\ \bottomrule
\end{tabular}
\label{cad120}
\end{table}
\section{Interpretation and Visualization}
\begin{figure}[htb]
  \centering
\vspace{-3mm}
  \includegraphics[width=\linewidth]{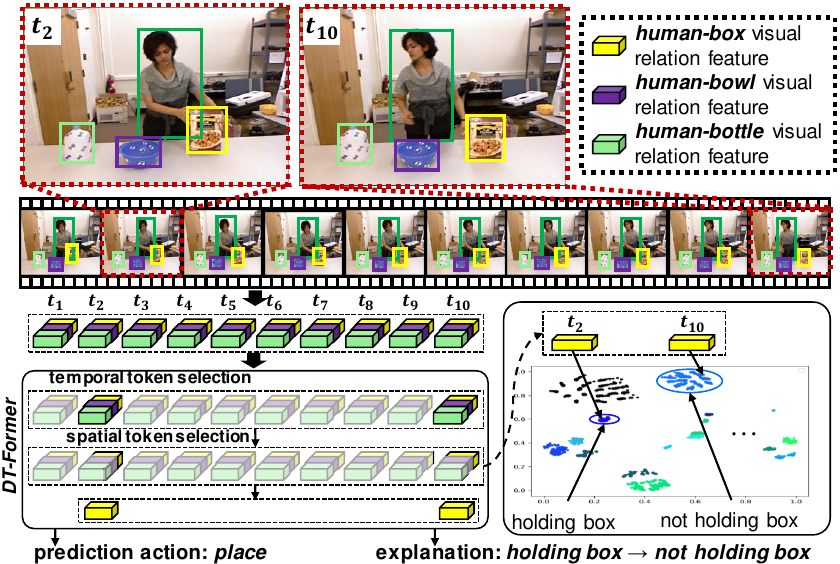}
   \caption{An example of action recognition performed by the proposed method and its corresponding process of providing explanations.
The shaded visual relation representations indicates that it is not selected by DT-Former.
}
   \label{fig:vis}
   \vspace*{-0.6cm}
\end{figure}

To intuitively demonstrate the interpretability effect of our proposed model, we provide interpretable representation and a visualization example.
As shown in Figure \ref{fig:vis}, in the inference stage, we first extract the visual relation representations of human-object pairs in each frame.
Then, our proposed DT-Former selects important relations in temporal and spatial dimensions and predicts action by modeling the important relation transition.
Finally, the visual representations of important relations are mapped into the joint embedding space to find their nearest neighbor semantic labels, which can provide explicit evidence for the action reasoning process.
In this example, the relation representations between the person and the box in the second and tenth frames are selected as cues for the action recognition.
The nearest semantic labels of these two representations in the joint embedding space are "holding box" and "not holding box", respectively.
Here, the consequences of observations \textbf{‘holding’} $\rightarrow$ \textbf{‘not holding’} provide a clear sign of the action "place". 

\section{Conclusion}

In this paper, we introduced a new framework, \textbf{LaIAR}, designed to transfer the knowledge from the language model to the video model to improve the recognition performance and interpretability of video models.
Specifically, we build a language model and a video model, which take semantic relation and visual relation representations as input, respectively.
These two models share the same architecture, namely DT-Former. 
This architecture is tailored to select the most important relations for action recognition from all the relations in video and to model the fine-grained relation transitions within videos.
Our framework also incorporates three novel knowledge transfer strategies in our learning scheme to facilitate the knowledge transfer from the language model to the video model.
This not only boosts the performance but also enhances the interpretability of the video model.
Ablation experiments verified the effectiveness of the DT-Former, the learning scheme module and the robustness against domain shift.
We conducted extensive experiments on Charades and CAD-120 datasets to demonstrate the superior performance of our proposed method.

\textbf{Acknowledgements:} This work is partially supported by the National Natural Science Foundation of China under Grant No.62073252 and No.62072358. It was also supported by Natural Science Basic Research Program of Shaanxi under Grant No.2024JC-JCQN-66.
{
    \small
    \bibliographystyle{ieeenat_fullname}
    \bibliography{main}
}


\end{document}